\documentclass[a4paper,fleqn]{cas-sc}
\usepackage[authoryear,longnamesfirst]{natbib}

\usepackage{wrapfig}
\usepackage{caption}
\usepackage{subcaption}
\usepackage{algorithmic}
\usepackage{graphicx}
\usepackage{xcolor}
\usepackage{color}
\usepackage{lscape} 
\usepackage{textcomp}
\usepackage{multirow}
\usepackage{url}          
\usepackage{booktabs}     
\usepackage{nicefrac}     
\usepackage{microtype}    
\usepackage{longtable}
\usepackage{pdflscape}
\usepackage{tabularray}
\usepackage{hyperref} 
\usepackage{float}
\usepackage{amsmath}
\usepackage{cleveref}
\DeclareMathAlphabet\mathbfcal{OMS}{cmsy}{b}{n}

\usepackage{bm}

\usepackage[most]{tcolorbox}
\tcbuselibrary{theorems}
\usepackage{cleveref}
\crefname{Teo}{mytheorem}{Mytheorem}
\newtcbtheorem[auto counter, crefname={}{Mytheorem}]{mytheorem}{Procedure}{colbacktitle=gray, coltitle=black}{Teo}

\crefname{pro}{myprompt}{Myprompt}
\newtcbtheorem[auto counter, crefname={}{Myprompt}]{myprompt}{Prompt}{colbacktitle=gray, coltitle=black}{pro}
\usepackage[bidi=default, english]{babel}
\usepackage[LAE, T1]{fontenc}

\begin{document} 
\let\WriteBookmarks\relax
\def\floatpagepagefraction{1}
\def\textpagefraction{.001}

\shorttitle{}

\shortauthors{A. Bahaj et~al.}

\title [mode = title]{ 
AsthmaBot: Multi-modal, Multi-Lingual Retrieval Augmented Generation For Asthma Patient Support
}
\author[1]{Adil Bahaj}
\cormark[1]
\ead{adil.bahaj@uir.ac.ma}


\credit{Conceptualization, Software, Formal analysis, Investigation, Methodology, Writing - original draft}

\affiliation[1]{organization={TICLab, College of Engineering \& Architecture, International University of Rabat},
    \city={Rabat},
   country={Morocco}}

\author[1, 2]{Mounir Ghogho}
\affiliation[2]{organization={Faculty of Engineering,University of Leeds},
   country={United Kingdom}}

\ead{mounir.ghogho@uir.ac.ma}
\credit{Conceptualization, Formal analysis, Project administration, Supervision, Validation, Writing - review \& editing}

\begin{abstract}
        Asthma rates have risen globally, driven by environmental and lifestyle factors. Access to immediate medical care is limited, particularly in developing countries, necessitating automated support systems. Large Language Models like ChatGPT (Chat Generative Pre-trained Transformer) and Gemini have advanced natural language processing in general and question answering in particular, however, they are prone to producing factually incorrect responses (i.e. hallucinations). Retrieval-augmented generation systems, integrating curated documents, can improve large language models' performance and reduce the incidence of hallucination. We introduce AsthmaBot, a multi-lingual, multi-modal retrieval-augmented generation system for asthma support. Evaluation of an asthma-related frequently asked questions dataset shows AsthmaBot's efficacy. AsthmaBot has an added interactive and intuitive interface that integrates different data modalities (text, images, videos) to make it accessible to the larger public. AsthmaBot is available online via \url{asthmabot.datanets.org}.
\end{abstract}




\maketitle

\section{Introduction}
    In the past few decades, asthma rates have been on the rise globally, attributed not just to genetic factors but primarily to the influence of numerous environmental and lifestyle risk factors \cite{nunes2017asthma}. Asthma claims thousands of lives every year mainly due to lack of access to immediate and adequate medical care \cite{mwangi2020iot}. However, a significant number of asthma-related fatalities are preventable by home remedies, exercise, treatments, and action plans which can help reduce the symptoms of asthma patients either by avoiding triggers or employing relieving remedies \cite{mwangi2020iot, murphy2021smartphone}. This shows the importance of easy and fast access to information in reducing the ramifications of asthma attacks. However, having an around-the-clock service by medical providers can be prohibitive in many ways, especially in developing countries, which motivates the need for an interactive automated medical support system for asthma patients.
    
    Large language models (LLMs) have garnered considerable attention in recent years due to their generative abilities \cite{brown2020language, chen2021evaluating, wei2021pretrained}. Models such as ChatGPT \cite{ouyang2022training}, Gemini \cite{team2023gemini}, and Llama2 \cite{touvron2023llama} have paved the way for a new era of artificial intelligence (AI) where humans can interact with models in a mutually inclusive way. These models revolutionized multiple aspects of natural language processing (NLP) tasks (information retrieval, question answering, summarization, sentiment analysis etc) \cite{ouyang2022training, team2023gemini, touvron2023llama}. Recently, a plethora of works showed the proficiency of LLMs in question answering. Although their performance is encouraging they were found to generate plausibly sounding but factually incorrect responses, commonly known as hallucinations. In addition, LLMs can only remember the data that they trained on hindering the recency of their knowledge \cite{huang2023survey}. These limitations can negatively affect critical domains such as healthcare.

    To address these challenges new works \cite{lewis2020retrieval, gao2023retrieval} supply LLMs with pertinent documents sourced from current and reliable collections. This technique is called retrieval augmented generation (RAG) and was first introduced in \cite{lewis2020retrieval}. RAG systems generally contain three main components: document collections (corpora), retrieval algorithms (retrievers), and backbone LLMs \cite{gao2023retrieval}. However, existing RAG systems only produce textual information and are not multi-modal in addition LLMs suffer from language biases, which limits the quality of information that they generate in languages other than English.

    In this work, we introduce AsthmaBot, an automated asthma medical support multi-lingual, multi-modal RAG system. It is designed to provide asthma patients with answers to their questions based on a recent and curated list of documents, videos and images that are relevant to their queries. We evaluated AsthmaBot on a list of asthma-related question-answer pairs based on frequently asked questions (FAQs) lists from multiple online sources. Although this work focuses on asthma our methodology can be applied to various domains (e.g. law, other diseases and medical concepts etc). In fact, the same system can be used for any set of text, image and video data without any additional steps. Our experiments show an increased performance relative to a no RAG baseline in different languages on different modalities. This work makes contributions on the application and methodology levels the following summarizes our contributions:
    \begin{itemize}
        \item Application:
        \begin{itemize}
            \item Creating an automated asthma medical support system which contains text and media (images \& videos).
            \item Evaluating our system on asthma question answering.
            \item Making the system accessible to the wider community via a web-based interface.
        \end{itemize}
        \item Methodology:
        \begin{itemize}
            \item Design and implementation of a multi-modal and multi-lingual RAG system.
            \item Proposing an automated evaluation process for a multi-modal, multi-language setting based on frequently asked questions (FAQs) and question-answer pairs.
            \item Empirically Proving the linguistic biases of the used LLM, which motivates the translation of queries to English.
            \item Proposing a visual interface that integrates the different modalities (text, images, and videos) and their sources in each AsthmaBot response.
        \end{itemize}
    \end{itemize}

\section{Previous Literature}
    \subsection{Large Language Models (LLMs)}
        Traditionally, language models (LM) were developed to model the sequential nature of text \cite{merity2018regularizing}. These language models evolved from simple LSTM-based models \cite{merity2018regularizing, graves2012long} to transformer-based models such as BERT \cite{kenton2019bert}. In general, these models were pre-trained in a general-purpose unsupervised task (e.g. next token prediction, next sentence prediction) and then fine-tuned to a specific task using supervised training. More recently, the surge in computational resources and increased efficiency of training techniques have allowed for a surmountable increase in transformer models' parameters, which coupled with a significant amount of available data on the web have allowed for the creation of LLMs \cite{ouyang2022training, team2023gemini}. LLMs came with a new set of abilities that the traditional models never exhibited, like the emergence property, which allows LLMs to do tasks that they were not trained on. However, LLMs suffer from hallucinations. Hallucinations describe model outputs that are linguistically coherent but nonfactual \cite{huang2023survey}. To reduce the effects of these hallucinations retrieval augmented generation was proposed \cite{lewis2020retrieval, gao2023retrieval}. In this setting an LLM is provided with a query relevant context that is extracted from factual sources to augment the limited knowledge of the LLM and increase the likelihood of having a factual answer. RAG systems have the following components: document collections (corpora), retrieval algorithms (retrievers), and backbone LLMs \cite{gao2023retrieval}. The document collection is a collection of text documents from various sources that can be general or domain-specific. Retrieval algorithms are tasked with retrieving relevant documents given a query. The backbone LLM is used to generate an answer to the query given retrieved documents augmented prompt. Additionally, the previous RAG system is based on text-only documents, recent works explored multi-modal RAG systems, which can retrieve not just text but also images \cite{chen2022murag, zhao2023retrieving}. 
    \subsection{Biomedical LLMs and RAG}
        NLP researchers found fertile ground for innovation in Biomedical NLP due to its knowledge-intensive nature. Medicine has benefited from the development of NLP techniques since the inception of the field \cite{lee2020biobert, gu2021domain}. Recently, LLMs have demonstrated a considerable improvement in existing approaches to medical benchmarks in various tasks (named entity recognition, question answering, medical text summarization, medical diagnosis etc) \cite{chen2023meditron, chen2023extensive}. Although multiple LLMs trained on medical text have been proposed, they still suffer from the same hallucinations as the general domain LLMs. Consequently, RAG-based systems for medical question answering have been proposed \cite{xiong2024benchmarking, miao2024integrating, zakka2024almanac, wang2024jmlr}. For example, \cite{wang2024jmlr} showed that a RAG system with a 13B parameter backbone LLM can significantly outperform a 70B model (i.e. MediTron \cite{chen2023meditron}). However, these RAG systems are not multi-modal or multi-lingual.
        
\section{Methodology}
    AsthmaBot was conceptualised and designed to allow asthma patients to access relevant information interactively and adaptively. AsthmaBot backend is a multi-modal, multi-lingual retrieval augmented generation LLM, while the frontend is in the form of a chatbot (e.g. ChatGPT, Gemini etc) where questions and answers are saved so that the user can explore them. 
    \subsection{Data Collections}
         To enrich LLM's responses with factual up-to-date information specific documents and media were selected. These elements satisfy certain criteria rather than just doing a web search every time a query is given by the user, which can leave room for pseudo-scientific sources. Our main data formats are PDF documents obtained by searching Google, images from Google images and videos from YouTube. Search for the elements is done via queries in the different search platforms. Queries can be general like "asthma" or more specific (e.g. FAQs). Table \ref{tab:queries} shows different queries used to obtain data. FAQs are obtained from the gina website\footnote{https://ginasthma.org/about-us/faqs/}. These FAQs were later translated into different languages to obtain data from languages other than English. The PDF documents were downloaded manually while images from Google Images and YouTube videos were scraped automatically. After scraping the different images and videos we extracted the text from the source of the images and the video transcripts of each image and video respectively. We should note that we explored using image captioning to obtain image description but it was limited and didn't give importance to the context in which the image was used, consequently, we opted for the text in the webpage of the image source as an alternative.
         

         
        \begin{table}[h]
            \centering
            \begin{tabular}{c|cc}
                \hline
                Data Type & Source & Query\\
                \hline
                 Documents & Google&"asthma", "asthme", "\foreignlanguage{arabic}{الربو}"\\
                 Images & Google Images&"asthma", "asthme", "\foreignlanguage{arabic}{الربو}", FAQs\\
                 Videos & YouTube&"asthma", "asthme", "\foreignlanguage{arabic}{الربو}", FAQs\\
                 \hline
            \end{tabular}
            \caption{Different queries for obtaining data from different modalities.}
            \label{tab:queries}
        \end{table}
        \subsection{Vector DB Building and Retrievers}
        A vector database indexes text, images, videos and other data modalities based on a textual description of what they contain. This textual description is transformed into a dense vector using a language model, which facilitates the process of semantic search. We built multiple independent vector databases where each containing different modalities in different languages. Although one database that contains all the aforementioned elements can be used, it causes certain limitations. First, having multiple stores makes parallelism possible, increasing the speed of the inference process. Second, the search process can be parameterized providing more structure and control to the LLM and reducing hallucinations.

        We built nine vector DBs for each modality (text, images, videos) in each language (French, Arab, English). For each modality, we obtain the indexing description by summarising document text, image source text and video transcripts for text, images and videos respectively. These descriptions are further translated into English if they are in French or Arabic. This choice was made: 1) to use high-performing embedding models in English, 2) to avoid the linguistic biases in LLMs which seem to favour English. We used the FAISS vector database \cite{douze2024faiss} to save and query the different elements. FAISS has a retriever that uses cosine similarity to retrieve passages that are close to the query.
        \subsection{Prompting and Backbone LLM}
        Prompting is a fundamental element of LLM inference as it has been shown that the quality and the design of the prompt affect the quality of the output of the LLM \cite{chen2023unleashing}. \cref{pro:rule_skill} shows the prompt that we used in AsthmaBot. The prompt contains three parameters the query, context and  "history". The query refers to the user input, the context refers to the different text passages obtained by querying the different text vector DBs. The "history" refers to the chat history between the user and the LLM, composed of question-answer pairs. We used the Google Gemini LLM to infer the results of different queries.
        \begin{myprompt}{Prompt for LLM inference}{rule_skill}
        INSTRUCTIONS:\\
        You are an asthma medical support provider called AsthmaBot. You are designed to be as helpful as possible while providing only factual information.\\
        You should be friendly, but not overly chatty. Context information is below.\\
        Given the context information and chat history and no prior knowledge, answer the query.
        Give a detailed answer.\\
        Your answer should encompass the whole context.
        
        CONTEXT:\\
          \{context\}
        
        CHAT HISTORY:\\
          \{history\}
        
        QUERY:\\
          \{question\}
        
        ANSWER:
    \end{myprompt}
    \subsection{Translation}
        Instead of opting for queries in the native language, we translate the queries to English, search the vector databases, prompt the model in English, and translate the model response to the native query language. This was done because we noticed that LLMs (Gemini and ChatGPT) have biases towards the English language and can produce significantly richer responses in English than in Arabic or French. We further illustrate this in the results section, where we query the LLM using Arabic and get a significantly worse result than if we query it using the described process. We opted for the Google Translate API since it gave better results and a faster response time compared to other translators.
    \subsection{Visual Interface}
\begin{figure}[h]
            \centering
            \includegraphics[scale=0.8]{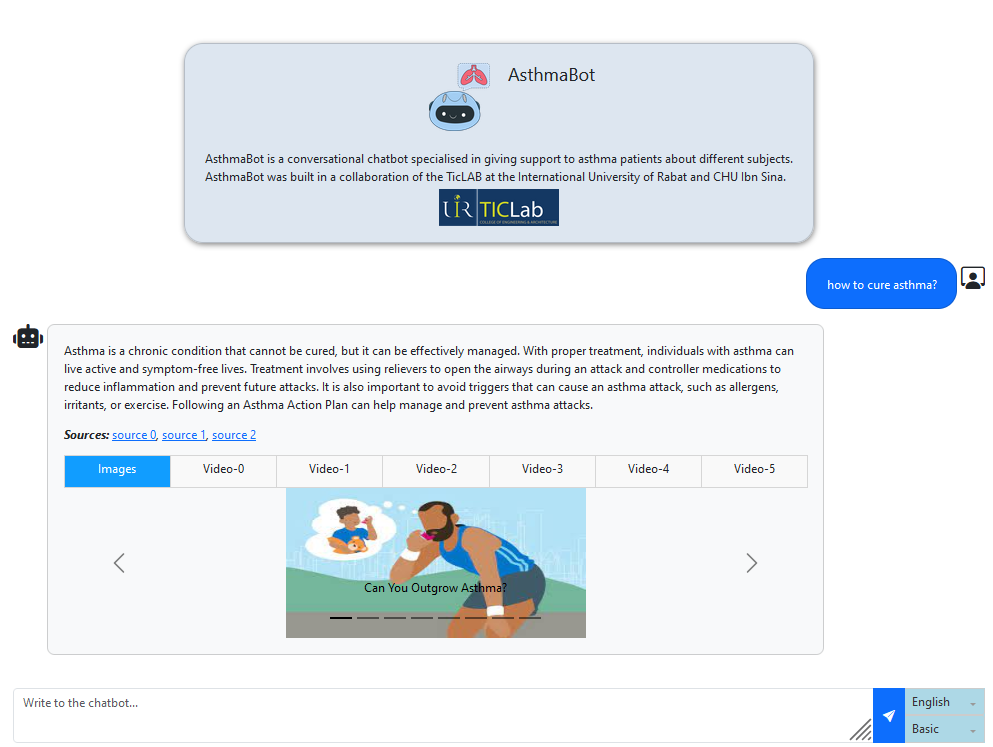}
            \caption{AsthmaBot interface.}
            \label{fig:asthmaBot_inter}
        \end{figure}

        To ensure the accessibility of AsthmaBot to a wider user base we created an interactive interface similar to that of ChatGPT and Gemini with the added features of embedded videos, source documents and images. In addition, users can click on the images to be transferred to the source website of the image. Figure \ref{fig:asthmaBot_inter} shows the interface of AsthmaBot.
    \subsection{AsthmaBot's Inference Process}
\begin{figure}
        \centering
        \includegraphics[scale=0.43]{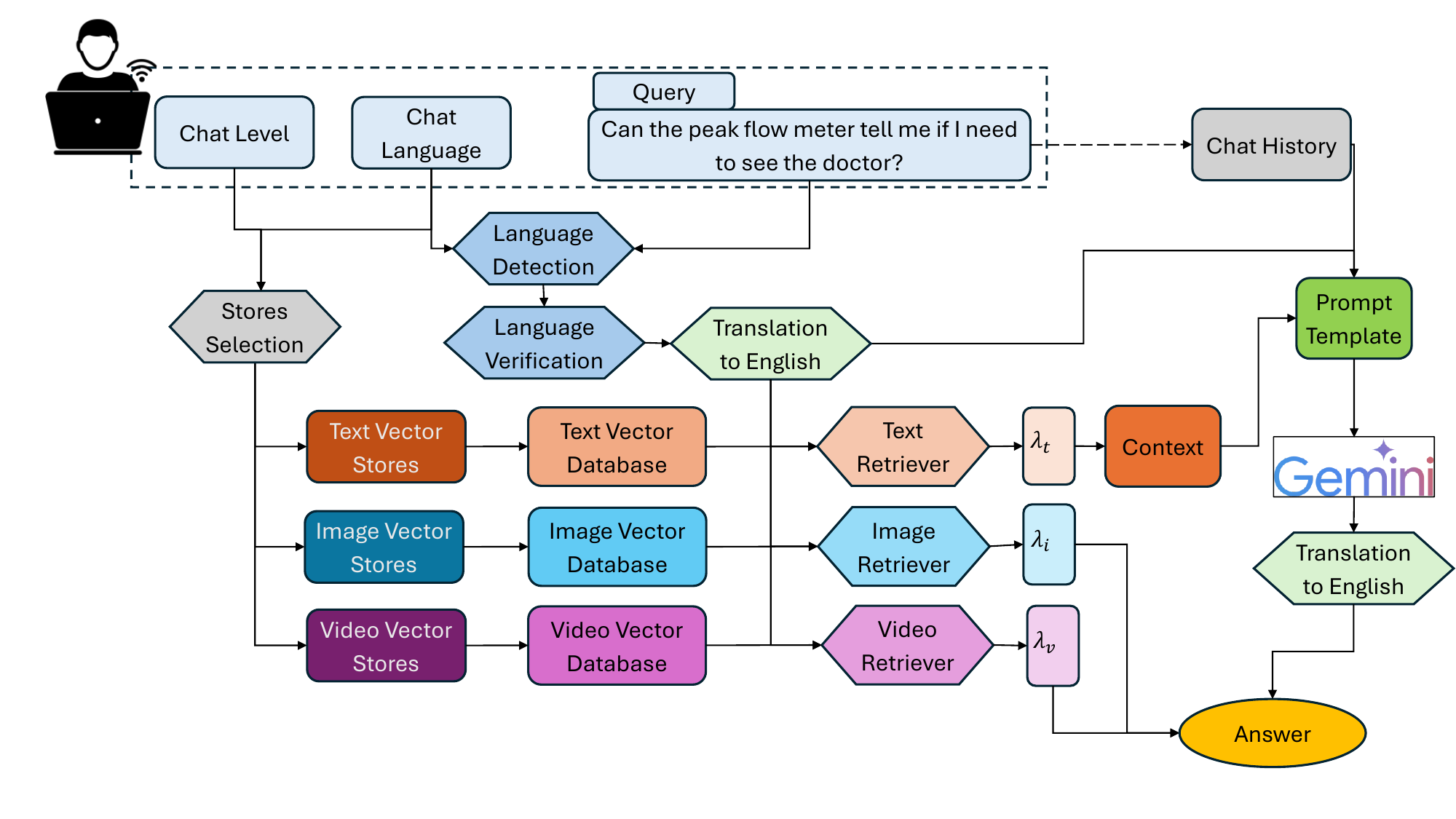}
        \caption{\textbf{AsthmaBot Overview}: given a query AsthmaBot uses asthma-related external resources to retrieve information relevant to the query and the query language before synthesizing a multi-modal response. This framework provides a multi-modal response grounded in truth, which reduces hallucinations and provides multiple formats for the answer.}
        \label{fig:system_process}
    \end{figure}

    AsthmaBots' inference process can be summarized as follows:
    \begin{itemize}
        \item Query: the process starts with a user inputting a query.
        \item Language detection: a language detection module detects the language of the query to choose the right vector stores.
        \item Query translation: if the detected language is not English then the query is translated to English.
        \item Vector store selection: AsthmaBot contains multiple vector stores: multiple text vector stores separated depending on language and content, and each one contains text chunks and their vector representation. Multiple image vector stores are separated depending on the language and indexed by the summary of the content of the image source document. Multiple video vector stores are separated depending on the language, where each video is indexed depending on a summary of its transcript. A vector store of each modality is selected.
        \item Retriever: a retriever searches the different vector stores to look for chunks of text, videos and images that are semantically similar to the query and where the similarity score is above a threshold. We have three thresholds for different text ($\lambda_{t}$), images ($\lambda_{i}$) and videos ($\lambda_{v}$) respectively.
        \item Prompt Generation: the retrieved text is fed to a prompt in addition to the query to augment the LLM response.
        \item Answer generation: the constructed prompt is fed to an LLM to generate the answer to the query.
        \item Answer translation: The output of the LLM is in English, consequently we have to translate it back to the detected language of the query.
    \end{itemize}
    Figure \ref{fig:system_process} shows the process of inference in AsthmaBot. This interface is available at \url{asthmabot.datanets.org}.

\section{Results}
    \subsection{Evaluation Procedure}
        To evaluate the efficacy of AsthmaBot at answering asthma-related questions we extracted a list of frequently asked questions (FAQs) and their corresponding answers from different sources and compared the answers generated by AsthmaBot to the answers originally given to the FAQs. In addition, to evaluate the veracity of information in the videos and images we modify the AsthmaBot context to video transcript summary and image source summary, respectively. In this way, we can automatically evaluate AsthmaBot's multi-modal retrieval capabilities without human intervention. We employed two automated evaluation metrics: ROUGE \cite{chin2004rouge} and BLEU \cite{papineni2002bleu}.
    \subsection{Evaluation Data}
        To evaluate AsthmaBot we collected a set of asthma-related FAQs from various sources. We collected over 400 FAQs from over 30 sources. Table \ref{tab:qa_pair_examples} gives various examples of question-answer pairs that we collected. To evaluate AsthmaBot's multilingual capabilities we translated the FAQs using Google Translate API.
        \begin{table}[]
            \centering
            \begin{tabular}{p{10em}|p{25em}}
                \hline
                Question & Answer \\
                \hline
                 How do I know when my child is old enough not to use a spacer?& Anyone using a metered-dose inhaler (MDI aka 'puffer') should always use a spacer.  A spacer helps more medicine reach your lungs.  If you don't want to use a spacer, you should talk to your doctor about getting a different asthma device.  Dry powder inhalers do not require spacers.  You should ask your pharmacist, your doctor, or your asthma educator to review your device technique every time you see them. It is harder to be good at taking your medicine than you might think. Consider using the same device for all your asthma medications so that you can become really good at one technique. \\
                 \hline
                 Why does asthma seem worse at night? & We have natural hormones (glucocorticoids) in our bodies that help keep the airways open by reducing inflammation. At nighttime when you are sleeping, these hormones are normally at lower levels, allowing more inflammation in the airways and increasing asthma symptoms. If your child has regular asthma symptoms at night, it may mean their asthma is poorly controlled and you should make an appointment to discuss this with your doctor. \\
                 \hline
                 Will my child become resistant to asthma medicine?& No. If asthma medicines become less effective, contact a healthcare provider for advice. Your health care provider may increase or change your medicines.\\
                 \hline
                 Why does my asthma get worse when I am upset or worried about something?& Not all people with asthma feel worse when they are upset or worried. Those who do may be easily stressed, or may cry or breathe too fast (hyperventilate) easily. Another reason that your asthma gets worse could be that you are not being treated properly for the inflammation you have in your airways.\\
                 \hline
                 Can asthma medication help prevent asthma symptoms?&Yes. Asthma medications include very effective airway openers. Even more importantly, they include very effective controllers (inhaled steroids), which can prevent most asthma attacks when used regularly.\\
                 \hline
            \end{tabular}
            \caption{Question-answer pair examples from the FAQs dataset.}
            \label{tab:qa_pair_examples}
        \end{table}
        
    \subsection{Evaluation Results}
        Table \ref{tab:performance_all} shows the results of querying using AsthmaBot in multiple languages (English, Arabic, French) in multiple data modalities (text, images, videos). The table shows that RAG significantly improves the performance in question answering relative to the no RAG baseline. This is true for all modalities and languages. On the other hand, we notice that there are variations in performance between languages. This can be attributed to the richness of the documents that were used in RAG.

        Table \ref{tab:nqvstq} shows the results of experimenting with using English-only input to the LLM and using prompts in the native language of the query. The results show that the English-only prompts perform significantly better than the native language prompt. This can be attributed to the significant amount of English prompts that the LLM was trained on. This language bias limits the richness of LLM output in languages other than English.
        \begin{table}[h]
            \centering
            \begin{tabular}{cc|cccc}
                \multicolumn{2}{c|}{Setting} & ROUGE-1 & ROUGE-2 & ROUGE-L & BLEU\\
                \hline
                \multirow{4}{*}{English} &No RAG&0.2370& 0.0504&0.1338&0.0180\\
                &Text&0.2684&0.0612& 0.1576&0.0367\\
                &Image &0.2664& 0.0585&0.1547&0.0327\\
                &Video&0.2686&0.0570&0.1556&0.0321\\
                \hline
                \multirow{4}{*}{French}&No RAG&0.2810& 0.0868&0.1528&0.0219\\
                &Text&0.2963&0.0952&0.1706&0.0426\\
                &Image &0.3108&0.1003& 0.1709&0.0441\\
                &Video&0.3024&0.0973&0.1696& 0.0408\\
                \hline
                \multirow{4}{*}{Arabic}&No RAG&0.0148&0.0056& 0.0149&0.0125\\
                &Text &0.0152& 0.0068&0.0149& 0.0300\\
                &Image &0.0162&0.0044&0.0162&0.0282\\
                &Video & 0.0169&0.0050&0.0167&0.0263\\
                \hline
            \end{tabular}
            \caption{Performance Evaluation in different languages for different modalities.}
            \label{tab:performance_all}
        \end{table}

        \begin{table}[h]
            \centering
            \begin{tabular}{cc|cccc}
                \multicolumn{2}{c|}{Setting} & ROUGE-1 & ROUGE-2 & ROUGE-L & BLEU\\
                \hline
                \multirow{3}{*}{Arabic} & No RAG &0.0148&0.0056& 0.0148&0.0125\\
                &RAG NQ&0.0106&0.0026&0.0106& 0.0176\\
                 &RAG TQ&0.0152& 0.0068&0.0149& 0.0300\\
                 \hline
                 \multirow{3}{*}{French} & No RAG&0.2810&0.0868&0.1528&0.0219\\
                 &RAG NQ&0.2963&0.0952&0.1706&0.0426\\
                 &RAG TQ&0.3181&0.0989&0.1761&0.0394\\
                 \hline
            \end{tabular}
            \caption{Results of querying with native language and with English translations of the query and the prompts. We conducted the same experiment for French and Arabic. NQ refers to "native query" and TQ refers to "translated query".}
            \label{tab:nqvstq}
        \end{table}
\section{Impact and Future Work}
    AsthmaBot integrates multiple features that are not included in many publicly available LLMs. We compared AsthmaBot to multiple publically available LLMs: ChatGPT\footnote{chat.openai.com}, Gemini \footnote{gemini.google.com}, Perplexity AI\footnote{perplexity.ai}, YouChat\footnote{you.com}. Table \ref{tab:llm_comp} compares the retrieval capabilities of different LLMs and AsthmaBot. Table \ref{tab:out} compares different LLMs generative capabilities. Although these models are made for general-purpose use, there is an increasing need for more domain-specific models with more constraints to reduce hallucination and language bias and AsthmaBot is another step in this direction for biomedical applications.
    
    \begin{table}[h]
        \centering
        \begin{tabular}{c|cccccc}
            LLM & \multicolumn{2}{c}{Document Retrieval}& \multicolumn{2}{c}{Image Retrieval}& \multicolumn{2}{c}{Video Retrieval}\\
            \hline
            & ML& SL& ML& SL& ML& SL\\
            \hline
            ChatGPT& No& No& No& No& No& No\\
            Gemini& Yes& Yes& No& Yes& No& No\\
            Perplexity AI& No& Yes& No& Yes& No& Yes\\
            YouChat& No& Yes& No& No& No&No\\
            \hline
            AsthmaBot& Yes& Yes& Yes& Yes& Yes& Yes\\
            \hline
        \end{tabular}
        \caption{Comparison of retrieval capabilities of different publicly available LLMs and AsthmaBot. "ML" refers to multi-lingual and "SL" refers to simple language.}
        \label{tab:llm_comp}
    \end{table}
    
    \begin{table}[h]
        \centering
        \begin{tabular}{c|cccc}
            & Language answer consistency&	Answer specificity&	Hallucinations\\
            \hline
            ChatGPT&	No&	No&	Yes\\
            Gemini&	No&	No&	Yes\\
            Perplexity AI&	No&	No&	No\\
            YouChat&	Yes&	No&	No\\
            AsthmaBot&	Yes&	Yes&	No\\
            \hline
        \end{tabular}
        \caption{Comparison of the outputs of different LLMs and AsthmaBot.}
        \label{tab:out}
    \end{table}
    There are multiple improvements that we think can further enhance the quality of AsthmaBot responses. The following are a few:
    \begin{itemize}
        \item Improving data curation: AsthmaBot RAG video and image modules are less curated than the text module that we took from educational resources. Consequently, improving having expert-curated videos and images can further improve the trustworthiness of AsthmaBot.
        \item Visual summary (infographics) generation: since AsthmaBot will be mainly used by patients with varying educational levels, visual summaries of LLM results can improve accessibility significantly especially for children.
        \item Automatic prompting: AsthmaBot can benefit from the new research in learnable prompts \cite{khattab2023dspy}. This is particularly important since AsthmaBot since its primarily destined for asthma patients. Consequently, fine-tuning prompts to FAQs can improve it.
        \item Extending to other diseases: Although AsthmaBot focuses mainly on asthma it can be customised to fit other diseases by adding more documents, images and videos and changing the prompt.
        \item Evaluation: In this work, we attempted a fully automated evaluation protocol, which may not convey the full extent of evaluation like factuality and adherence to contextual information. On the other hand, these aspects can only be evaluated manually for now, which is prohibitive in our case.
    \end{itemize}
\section{Conclusion}

In this paper, we have presented AsthmaBot, a multi-lingual, multi-modal Retrieval-Augmented Generation (RAG) system designed to provide automated support for asthma patients. Through the integration of curated documents, videos, and images, AsthmaBot offers personalized responses to asthma-related queries, empowering patients with valuable information. Our evaluation, based on diverse asthma-related FAQs, demonstrates AsthmaBot's enhanced performance compared to a non-RAG baseline, highlighting its effectiveness in providing relevant information. Moreover, AsthmaBot tackles the language biases inherent in Large Language Models (LLMs) by offering multi-lingual support. The visual interface of AsthmaBot further enhances user experience, presenting information in a comprehensive and accessible manner. Although this work focuses on applying the system to asthma it can be applied to any domain containing text, video and image information (e.g. law, entertainment, other medical conditions and concepts etc).



\bibliographystyle{elsarticle-harv} 
\bibliography{ref}

\end{document}